\documentclass[10pt,twocolumn,letterpaper]{article}

\usepackage{iccv}
\usepackage{times}
\usepackage{epsfig}
\usepackage{graphicx}
\usepackage{amsmath}
\usepackage{amssymb}
\usepackage{bm}
\newcommand{\minus}{\scalebox{0.75}[1.0]{$-$}}
\usepackage{caption}
\usepackage{subcaption}
\usepackage{booktabs}
\usepackage{multirow}
\usepackage{multicol}
\newcommand{\tabincell}[2]{\begin{tabular}{@{}#1@{}}#2\end{tabular}}  
\usepackage[pagebackref=true,breaklinks=true,letterpaper=true,colorlinks,bookmarks=false]{hyperref}
\newcommand{\ding}[1]{\textcolor{black}{#1}} 
\newcommand{\xin}[1]{\textcolor{black}{#1}}
\usepackage[pagebackref=true,breaklinks=true,letterpaper=true,colorlinks,bookmarks=false]{hyperref}

\iccvfinalcopy 

\def\iccvPaperID{2963} 

\ificcvfinal\fi

\begin{document}

\title{M2IOSR: Maximal Mutual Information Open Set Recognition}

\author{
Xin Sun$^{1}$\footnotemark[1]
\quad
Henghui Ding$^{1,2}$\footnotemark[2]
\quad
Chi Zhang$^1$
\quad
Guosheng Lin$^1$
\quad
Keck-Voon Ling$^1$
\\
$^1$Nanyang Technological University, Singapore\qquad $^2$ByteDance
\\
{\tt\small \{xin001, ding0093, chi007, gslin, ekvling\}@ntu.edu.sg}
}
\maketitle
\renewcommand{\thefootnote}{\fnsymbol{footnote}}
\footnotetext[1]{Work partly done when Xin Sun was an intern at ByteDance AI Lab.}
\footnotetext[2]{Henghui Ding is the corresponding author.}

\maketitle
\ificcvfinal\thispagestyle{empty}\fi

\begin{abstract}
In this work, we aim to address the challenging task of open set recognition (OSR). Many recent OSR methods rely on auto-encoders to extract class-specific features by a reconstruction strategy, requiring the network to restore the input image on pixel-level. This strategy is commonly over-demanding for OSR since class-specific features are generally contained in target objects, not in all pixels. To address this shortcoming, here we discard the pixel-level reconstruction strategy and pay more attention to improving the effectiveness of class-specific feature extraction. We propose a mutual information-based method with a streamlined architecture, Maximal Mutual Information Open Set Recognition (M2IOSR). The proposed M2IOSR only uses an encoder to extract class-specific features by maximizing the mutual information between the given input and its latent features across multiple scales. Meanwhile, to further reduce the open space risk, latent features are constrained to class conditional Gaussian distributions by a KL-divergence loss function. In this way, a strong function is learned to prevent the network from mapping different observations to similar latent features and help the network extract class-specific features with desired statistical characteristics. The proposed method significantly improves the performance of baselines and achieves new state-of-the-art results on several benchmarks consistently.
\end{abstract}
\section{Introduction}
\begin{figure}[t]
\centering
\begin{tabular}{c@{\hspace{0.45mm}}c@{\hspace{0.45mm}}c}
{\small Image} & {\small Auto-encoders} & {\small Ours}\\
\includegraphics[width=0.31\linewidth, height=0.24\linewidth]{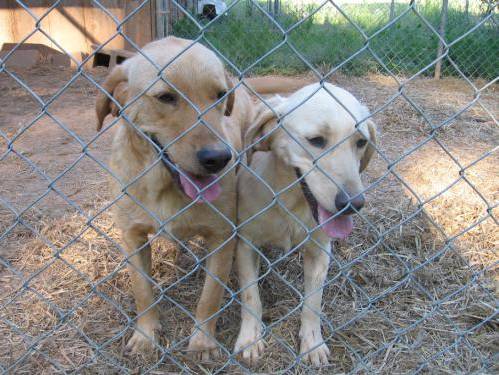}&
\includegraphics[width=0.31\linewidth, height=0.24\linewidth]{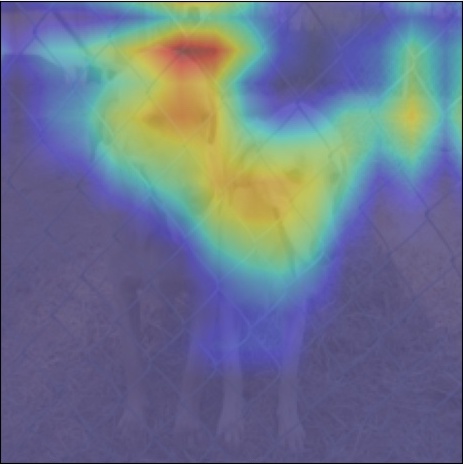}&
\includegraphics[width=0.31\linewidth, height=0.24\linewidth]{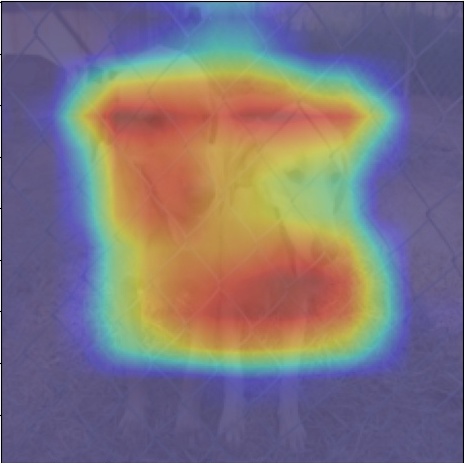}
\vspace{-0.5mm}\\
\includegraphics[width=0.31\linewidth, height=0.24\linewidth]{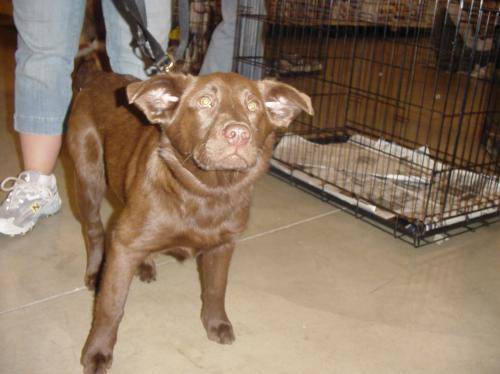}&
\includegraphics[width=0.31\linewidth, height=0.24\linewidth]{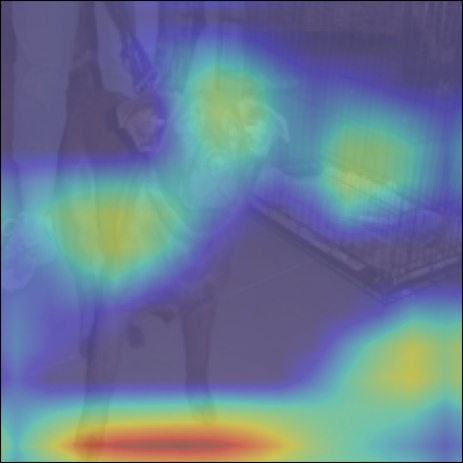}&
\includegraphics[width=0.31\linewidth, height=0.24\linewidth]{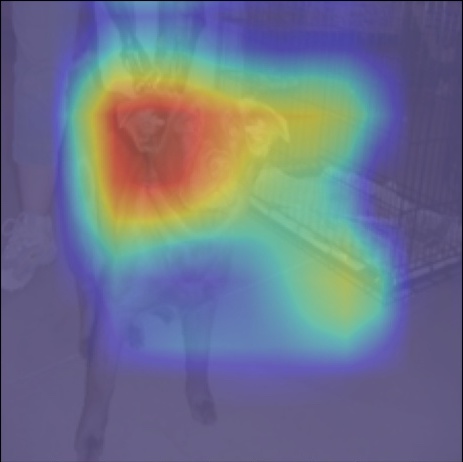}
\vspace{-0.5mm}\\
\includegraphics[width=0.31\linewidth, height=0.24\linewidth]{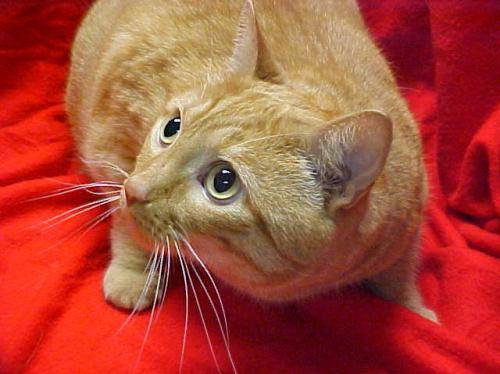}&
\includegraphics[width=0.31\linewidth, height=0.24\linewidth]{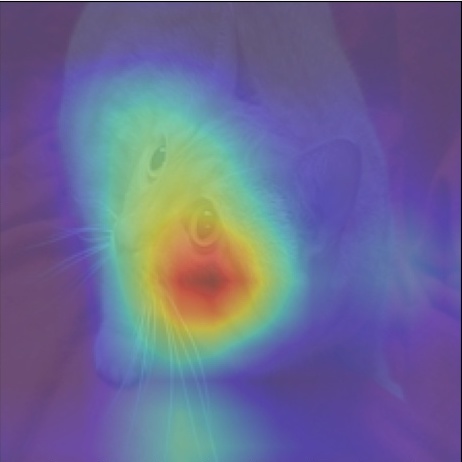}&
\includegraphics[width=0.31\linewidth, height=0.24\linewidth]{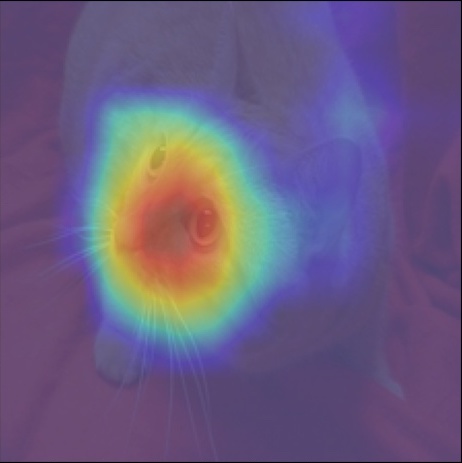}
\vspace{-0.5mm}\\
\end{tabular}
\caption{Class activation maps produced by auto-encoders and the proposed method. Due to the reconstruction of all pixels including target objects and background stuff, auto-encoders make the features of target objects less pronounced (\eg, the first two images). By contrast, the proposed method can extract more class-specific features, and thus its extracted features focus more on target objects. When the background ratio is small, \eg, the third image, our method localizes the target object more accurately than auto-encoders.}
\label{heatmap}
\end{figure}

Deep learning-based methods have demonstrated remarkable success in many recognition tasks, where deep neural networks are trained with a finite set of labeled training samples from $K$ distinct known classes~\cite{resnet, vgg, dense, ding2018context, ding2019boundary,ding2019semantic,dingiccv2021}. During inference, these networks trained on pre-defined $K$ classes typically select a label from these $K$ classes and assign the selected label to the input. Nevertheless, in most real-world classification tasks, the number of classes is usually much larger than $K$. Due to the limitations of various objective factors, it is commonly hard to build a dataset that covers all classes in the real-world. 
As a result, when a sample outside the known class set, \ie, an unknown sample, is fed into the network during testing, the network will make misclassification and incorrectly recognize this sample as one of the known $K$ classes. A potential solution to address this issue is open set recognition (OSR)~\cite{toward}. \ding{OSR assumes that deficient knowledge exists during the training stage, and aims to not only classify samples of known classes but detect unknown samples during the testing stage.}

To train an effective OSR classifier, the features fed into the classifier should be specific enough since only samples from known classes are available at training time. With class-specific features, the classifier can categorize known classes as well as identify unknown samples during inference~\cite{gdfr}. Many recent OSR methods~\cite{gdfr,cgdl,crosr,c2ae} rely on auto-encoders (AE)~\cite{ae} to extract class-specific features.
AE models are well known to extract class-specific features through a pixel-level reconstruction strategy, where each pixel in the reconstructed image is expected to be exactly the same as the input image. It is usually resource-consuming due to the decoder and complex network architecture. Besides, as shown in Fig.~\ref{heatmap}, the class activation maps indicate that an AE model weakens the feature saliency of target objects and only locates the target object well when the proportion of background is small.
These limitations have rendered that such a pixel-level reconstruction strategy is less suitable for OSR tasks. As class-specific features are generally contained in target objects, not in the background, there is no need to extract features by pixel-level. It is experimentally observed in Fig.~\ref{heatmap} that a non-pixel-level model is more accurate in class-specific feature extraction and robust in target localization than a pixel-level reconstruction model (see Sec.~\ref{case_sec} for experiment details).

Considering the above discussions, we discard the pixel-level reconstruction strategy and pay more attention to improving the effectiveness of class-specific feature extraction. A novel OSR method, Maximal Mutual Information Open Set Recognition (M2IOSR), is thus proposed in this work.
Different from most previous deep-learning-based OSR methods~\cite{g-openmax,counter,crosr,c2ae,cgdl,gdfr}, M2IOSR has a streamlined architecture and only uses an encoder to discover class-specific features by maximizing the mutual information (MI)~\cite{shannon} between the input and its latent features.
\ding{The mutual information greatly helps the encoder establish stronger interdependence between the input image and its learned features for OSR.}

Following the set up in~\cite{dim}, each training iteration of the proposed method includes two steps: maximizing MI on positive pairs from the same input and minimizing MI on negative pairs from different inputs. For positive/negative pairs, the proposed method maximizes/minimizes the MI between the complete input and latent features (global MI), as well as the average MI between every {local} region (patches) of the input and latent features (local MI). Meanwhile, to further reduce the open space risk~\cite{toward}, it also encourages latent features to have desired statistical characteristics. For statistical constraint, we replace the adversarial learning in~\cite{dim} by the KL-divergence loss function proposed in~\cite{cgdl}. At the same time, inspired by~\cite{amdim}, to prevent MI from being lost in the middle layers, we also maximize/minimize the local MI across multiple scales. In this way, a strong function can be learned to avoid the network from mapping different observations to similar latent features and help the network extract class-specific features with class conditional Gaussian distributions.

In experiments, we carefully study and demonstrate the effectiveness of MI maximization strategy, KL-divergence loss function, multi-scaled local MI, and positive/negative pairs, on improving OSR performance. The contributions of this paper are listed in the following:
\begin{itemize}
\setlength
  \item We propose a novel OSR method, Maximal Mutual Information Ope Set Recognition (M2IOSR). Different from most previous deep learning-based OSR methods, the proposed method only uses an encoder architecture to extract class-specific features for OSR.
  \item To further reduce the open space risk, we adopt a KL-divergence loss function to encourage latent features to follow class conditional Gaussian distributions. 
  \item To prevent MI from being lost in the middle layers, we incorporate multi-scaled local MI maximization.
  \item Experiments on several benchmarks demonstrate that the proposed method significantly enhances the performance of our baseline and achieves new state-of-the-art OSR performance.
\end{itemize}
\section{Related Work}

\textbf{Open Set Recognition.}
\xin{In~\cite{toward}, the concept of {Open Set Recognition (OSR)} was put forward and an SVM-based OSR method (1-vs-Set machine) was introduced, which added an extra hyper-line to separate unknown samples from known samples. WSVM~\cite{wsvm} and $P_I$-SVM~\cite{pisvm} improved this 1-vs-Set machine by introducing the extreme value theory into the SVM to further control the open space risk. SROSR~\cite{sparse} emphasized on extracting useful information from reconstruction error distributions. Some distance-based OSR methods were also proposed. NNO~\cite{openworld} recognized those samples away from the centroids of the known classes as unknown. NNDR~\cite{nearest} classified known classes and detected unknown samples by evaluating similarity scores between the two most similar classes.}

The above methods are mostly based on traditional machine learning techniques (\eg, support vector machine, sparse representation, nearest neighbor, etc.). With deep learning achieving outstanding results in many computer vision fields, in recent years, more and more researchers focus on using deep learning techniques to address OSR tasks. One straightforward OSR method is setting a threshold on the probability of the most probable class produced by the Softmax layer~\cite{baseline}, but unknown samples could produce equally high probabilities~\cite{toward}. Openmax~\cite{openmax} overcame this shortcoming by redistributing the probability distribution produced from the Softmax function to get the predicted probability of the unknown class. Differently, some works directly estimated probabilities of unknown samples by adding synthesized unknown-class images into the training set~\cite{zhanghui2021}, such as G-Openmax~\cite{g-openmax} and OSRCI~\cite{counter}. Besides, a hybrid model proposed in~\cite{hybrid} combined an AE model, a classifier, and a density estimator to detect unknowns. RPL~\cite{reciprocal} classified known and unknown samples by the embedding of other samples with reciprocal points. Auto-encoders (AE) are also widely used in OSR tasks. CROSR~\cite{crosr} combined predicted scores and latent features to get probabilities of unknown samples. C2AE~\cite{c2ae} determined the decision boundary by using the extreme value theory on reconstruction errors. CGDL~\cite{cgdl} proposed a new KL-divergence loss function which makes learned features follow class conditional Gaussian distributions. GDFR~\cite{gdfr} extracted optimal features by feeding the input and its reconstruction into a self-supervised learning model. 

\textbf{Out-Of-Distribution (OOD)} detection~\cite{ood_iclr,ood_cvpr,ood_nips} is a similar task with OSR, but OOD more focuses on unknown detection, where known and unknown samples are usually from different domains. OSR needs to classify known classes and additionally reject unknown samples, where known and unknown samples can come from the same domain.

\xin{\textbf{Mutual Information Estimation.}}
{Mutual information (MI)} in unsupervised learning has a long history~\cite{1988mi,1992mi,1995mi,1996mi} as prescribed for neural networks. Recently, MI is widely used in deep neural networks for unsupervised discrete representation learning~\cite{imsat}, scene representation learning~\cite{scene}, segmentation and clustering~\cite{mi-se}. Especially, MINE~\cite{mine} learned an MI estimate of continuous variables based on generative models. DIM~\cite{dim} followed MINE in this respect, but emphasized the importance of maximizing local MI, which was also mentioned in other literature~\cite{amdim,cpc}. Some researchers also study on finding better bounds on MI used in deep learning~\cite{ mi-lim,mi-bound}.
\section{Preliminaries}
\label{preli}
Before describing our proposed approach, we here briefly revisit the theory of mutual information \cite{shannon}.

In information theory, the Mutual Information (MI) $I(A; B)$ is a measure of the mutual dependence between the two variables $A$ and $B$. More specifically, it quantifies the ``amount of information'' about the variable $A$ (or $B$) captured via observing the other variable $B$ (or $A$). The MI can be expressed as the difference between two entropy terms:
\begin{equation}
I(A;B)=H(A)-H(A|B)=H(B)-H(B|A),
\end{equation}
where $H(A)$ and $H(B)$ are the marginal entropy of $A$ and $B$. This definition has an intuitive interpretation: $I(A; B)$ is the reduction of uncertainty in one variable when the other variable is observed. 
\ding{If $A$ and $B$ are independent to each other, their MI is zero, \ie $I(A; B)=0$. In this case, knowing $A$ does not give any information about $B$ and vice versa. At the other extreme, if $A$ and $B$ are interdependent by a deterministic function, then all information conveyed by $A$ is shared with $B$. Thus, knowing $A$ determines the value of $B$ and vice versa, where the maximum MI is attained.}

In the case of jointly continuous random variables, the MI can be expressed as the following formulation:

\begin{equation}
\begin{aligned}
I(A;B)&=\int\int p(a,b)log\frac{p(a,b)}{p(a)p(b)}dadb\\
&=\int\int p(b|a)p(a)log\frac{p(b|a)}{p(b)}dadb,
\end{aligned}
\end{equation}
where $p(a,b)$ is the joint probability density function of $A$ and $B$, and $p(a)$ and $p(b)$ are the marginal probability density functions of $A$ and $B$ respectively.

\section{Proposed Method}
\ding{In this section, firstly, we introduce the derivation of loss function used in the proposed approach, Maximal Mutual Information Open Set Recognition (M2IOSR). Then we demonstrate the details of our MI maximization strategy. Finally, we present our training and testing procedures.}

\subsection{Loss Function Derivation}
The MI $I(\bm{X};\bm{Z})$ measures how much the mutual dependence between the input image $\bm{X}$ and the corresponding learned feature $\bm{Z}$. Maximizing MI $I(\bm{X};\bm{Z})$ ensures the learned feature $\bm{Z}$ is highly semantic and representative to describe the input image, 
\ding{which is desired for detecting unknown samples and assigning correct labels to known samples~\cite{gdfr}.}
As introduced in Sec.~\ref{preli}, the MI $I(\bm{X};\bm{Z})$ can be expressed as the following formulation:

\begin{equation}
I(\bm{X};\bm{Z})=\int\int p(\bm{z}|\bm{x})p(\bm{x})\log\frac{p(\bm{z}|\bm{x})}{p(\bm{z})}d\bm{x}d\bm{z},
\label{ixz}
\end{equation}
where \ding{$p(\bm{x},\bm{z}) = p(\bm{z}|\bm{x})p(\bm{x})$} is the joint probability density function of $\bm{X}$ and $\bm{Z}$, $p(\bm{x})$ and $p(\bm{z})$ are the marginal probability density functions of $\bm{X}$ and $\bm{Z}$ respectively. 

To further reduce open space risk~\cite{toward}, 
\xin{latent features should be constrained to some statistical property (\eg, Gaussian distribution).} Different from~\cite{dim} that employs adversarial learning, here we adopt a KL-divergence function to obtain desired latent features. Our training objective is to:~1) maximize MI $I(\bm{X};\bm{Z})$ between the input $\bm{X}$ and the learned feature $\bm{Z}$; and 2) minimize the KL-divergence between the posterior distribution $p(\bm{z})$ and the prior distribution $q(\bm{z})$. The loss function used in our M2IOSR can be expressed in the following:
\begin{equation}
L=\minus L_{MI}+\gamma L_{KL},
\label{loss_sum}
\end{equation}
where $\gamma$ is the balancing parameter\ding{, and $L_{MI}$ and $L_{KL}$ respectively denotes the MI loss and the KL-divergence loss}. 
The general form of MI is expressed in Eqn.~\ref{ixz}, and the general form of KL-divergence function is
\begin{equation}
KL\left[p(\bm{z})||q(\bm{z})\right]=\int p(\bm{z})\log\frac{p(\bm{z})}{q(\bm{z})}d\bm{z}.
\end{equation}
Herein, the training objective in Eqn.~\ref{loss_sum} can be expressed as the following formulation:
\begin{equation}
\begin{aligned}
\mathop{min}\limits_{p(\bm{z}|\bm{x})}\bigg\{&\minus\int\int p(\bm{z}|\bm{x})p(\bm{x})\log\frac{p(\bm{z}|\bm{x})}{p(\bm{z})}d\bm{x}d\bm{z}\\
&+\gamma\int p(\bm{z})\log\frac{p(\bm{z})}{q(\bm{z})}d\bm{z}\bigg\}.
\label{loss}
\end{aligned}
\end{equation}
As $p(\bm{z})=\int p(\bm{z}|\bm{x})p(\bm{x})d\bm{x}$, Eqn.~\ref{loss} is converted to:
\begin{equation}
\begin{aligned}
&\mathop{min}\limits_{p(\bm{z}|\bm{x})}\bigg\{\int\int p(\bm{z}|\bm{x})p(\bm{x})\Big[\minus(1+\gamma)\log\frac{p(\bm{z}|\bm{x})}{p(\bm{z})}\\
&+\gamma\log\frac{p(\bm{z}|\bm{x})}{q(\bm{z})}\Big]d\bm{x}d\bm{z}\bigg\}\\
&=\mathop{min}\limits_{p(\bm{z}|\bm{x})}\bigg\{\minus(1+\gamma) \Big[KL\big(p(\bm{z}|\bm{x})p(\bm{x})||p(\bm{z})p(\bm{x})\big)\Big]\\
&+\gamma \mathbb{E}_{\bm{x}\sim p(\bm{x})}\Big[KL\big(p(\bm{z}|\bm{x})||q(\bm{z})\big)\Big]\bigg\}.
\label{loss2}
\end{aligned}
\end{equation}
\ding{where $\mathbb{E}[*]$ denotes the mathematical expectation of variable $*$.} \xin{Considering that the KL-divergence function does not have an upper boundary, here we use the JS-divergence function to substitute the KL-divergence function in the first term~\cite{dim}. Following the Eqn.~6 and Tab.~6 in~\cite{fgan} and using the KL-divergence loss function proposed in~\cite{cgdl} as the second term, the complete training objective of M2IOSR is converted from Eqn.~\ref{loss2} to a more clear and implementable formulation:
\begin{equation}
\begin{aligned}
\mathop{min}\limits_{p(\bm{z}|\bm{x}),T(\bm{x},\bm{z})}\bigg\{&\minus(1+\gamma)\Big[E_{\bm{x}\sim p(\bm{z}|\bm{x})p(\bm{x})}\big(\log\sigma(T[\bm{x},\bm{z}])\big)\\
&+E_{\bm{x}\sim p(\bm{z})p(\bm{x})}\big(\log(1-\sigma(T[\bm{x},\bm{z}]))\big)\Big]\\
&+\gamma \mathbb{E}_{\bm{x}_k\sim p(\bm{x}_k)}\Big[KL\big(p(\bm{z}|\bm{x}_k)||q^{(k)}(\bm{z})\big)\Big]\bigg\},
\label{loss3}
\end{aligned}
\end{equation}
where $\sigma$ is an activation function and $T$ is a discriminator. The first term is the MI loss $L_{MI}$, which contains two parts: maximizing MI on positive pairs from the same input, and minimizing MI on negative pairs from different inputs. When $\bm{x}\sim p(\bm{z}|\bm{x})p(\bm{x})$, $[\bm{x},\bm{z}]$ is a positive pair as the latent feature $\bm{z}$ is extracted from its corresponding input $\bm{x}$. While when $\bm{x}\sim p(\bm{z})p(\bm{x})$, $[\bm{x},\bm{z}]$ is a negative pair as $\bm{z}$ is extracted from a random input. The second term is the KL-divergence loss $L_{KL}$. $p(\bm{z}|\bm{x}_k)$ is the $k$-th class conditional posterior distributions of latent features where $k$ is the index of known classes, and $q^{(k)}(\bm{z})$ is the $k$-th multivariate Gaussian distribution $N(\bm{z};\bm{\mu}_k,\bm{I})$.}

\begin{table*}[htbp]
\centering
\begin{minipage}[t]{0.48\linewidth}\centering
\caption{Global discriminator architecture.}
\label{global_dis}
\begin{tabular}{c c c}
\hline
Layer & Size & Activation \\
\hline
Linear Layer & 512 & ReLU\\
Linear Layer & 512 & ReLU\\
Linear Layer & 1 & Softplus\\
\hline
\end{tabular}

\end{minipage}\hfill%
\begin{minipage}[t]{0.48\linewidth}\centering
\caption{Local discriminator architecture.}
\label{local_dis}
\begin{tabular}{c c c}
\hline
Layer & Dimension & Activation \\
\hline
$1 \times 1$ Convolutional Layer & 512 & ReLU\\
$1 \times 1$ Convolutional Layer & 512 & ReLU\\
$1 \times 1$ Convolutional Layer & 1 & Softplus\\
\hline
\end{tabular}
\end{minipage}
\end{table*}

\begin{figure} [htbp]
    \centering
    \includegraphics[scale=0.39]{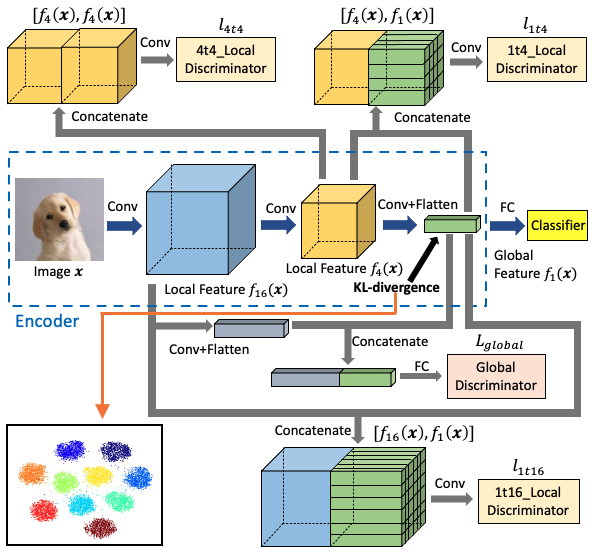}
    \caption{\xin{Overview architecture of the proposed method on positive pairs. $f_d(\bm{x})$ represents a feature with a spatial size of $d\times d$. During the training phase, we maximize global MI on $[f_{16}(\bm{x}), f_{1}(\bm{x})]$ to extract useful information of the input, and maximize local MI on $[f_{16}(\bm{x}), f_{1}(\bm{x})]$, $[f_4(\bm{x}), f_1(\bm{x})]$ and $[f_4(\bm{x}), f_4(\bm{x})]$ to filter out useless features and prevent MI from being lost in the middle layers. Meanwhile, to further reduce open space risk~\cite{toward}, latent features $f_1(\bm{x})$ are constrained to class conditional Gaussian distributions. A 2-dimensional tSNE~\cite{tsne} visualization of $f_1(\bm{x})$ distributions on the CIFAR10 dataset~\cite{cifar10} is shown at the bottom-left.}}
    \label{framework}
\end{figure}

\subsection{Mutual Information Maximization}
The overall architecture of the proposed method on positive pairs is shown in Fig.~\ref{framework}. To have a fair comparison with previous works, the backbone used here is the re-designed VGG-13 network defined in \cite{crosr}, and the dimension of latent features is fixed to 32. Here we use $f_d(\bm{x})$ to denote a feature with  
\ding{a spatial size of $d\times d$.}
Following the set-up in~\cite{dim}, the proposed method maximizes MI between the latent feature vector and the feature map after the first few layers (global MI) and maximizes average MI between the feature vector and every local region (patches) of the feature map (local MI), as well as encourages the representations to have desired characteristics. There are three layers of features are used in the proposed method, whose spatial size is respectively 16, 4, and 1, namely $f_{16}(\bm{x})$, $f_4(\bm{x})$, and $f_1(\bm{x})$.
The loss function used in M2IOSR is extended from Eqn.~\ref{loss_sum} to:
\begin{equation}
L=\minus (\beta_{1} L_{global}+\beta_{2} L_{local})+\gamma L_{KL}
\label{loss_sum2}
\end{equation}
where $\beta_1$, $\beta_2$ and $\gamma$ are the balancing parameters.

\textbf{Global MI}. \xin{Maximizing the MI between the whole input image and its corresponding latent feature helps the encoder extract useful information to pass through the network~\cite{dim}}. To this end, we encode the input $\bm{x}$ into the deep feature $f_{16}(\bm{x})$, 
\ding{whose spatial size is $16\times 16$.}
\ding{$f_{16}(\bm{x})$ is then processed by two convolutional layers with the kernel size of $3\times3$. The two layers' output is flattened as a vector to concatenate with the latent feature $f_1(\bm{x})$. The concatenated feature vector is input to a global discriminator $T_g$, which consists of two fully-connected layers, to get the {global} score. The activation function $\sigma_g$ used here is a {Softplus} function. Details of the global discriminator architecture are listed in Table~\ref{global_dis}. For positive pairs, $f_{16}(\bm{x})$ and $f_1(\bm{x})$ are from the same image. Negative pairs are formed by linking the latent feature $f_1(\bm{x})$ with a $f_{16}(\bm{\hat{x}})$ from another image.}


\textbf{Local MI.} \xin{Maximizing the average MI between local patches of the input and the latent features helps the encoder filter out noisy features~\cite{dim}}. For this objective, $f_1(\bm{x})$ is concatenated with the feature map $f_{16}(\bm{x})$ at every location. We use two $1 \times 1$ convolutional layers as the local discriminator to score the local MI pair $[f_{16}(\bm{x})$, $f_1(\bm{x})]$ ($l_{1t16}$). The activation function $\sigma_d$ used here is also a Softplus function. Details of the local discriminator architecture are listed in Table~\ref{local_dis}. 

\textbf{Multi-scaled Local MI.} \ding{To prevent MI from being lost in the middle layers, we form another two positive pairs $[f_{4}(\bm{x}),f_1(\bm{x})]$ ($l_{1t4}$), $[f_{4}(\bm{x}),f_4(\bm{x})]$ ($l_{4t4}$). Correspondingly, we also utilize three different negative pairs for Local MI during training: $[f_{16}(\bm{\hat{x}}), f_{1}(\bm{x})]$, $[f_{4}(\bm{\hat{x}}), f_{1}(\bm{x})]$, and $[f_{4}(\bm{\hat{x}}), f_{4}(\bm{x})]$, where $\bm{\hat{x}}$ and $\bm{x}$ are different images. The local MI loss $L_{local}$ in Eqn.~\ref{loss_sum2} is consisted of three terms:}
\begin{equation}
L_{local}=a_1l_{1t16}+a_2l_{1t4}+a_3l_{4t4}
\label{local_loss}
\end{equation}
where $a_1$, $a_2$ and $a_3$ are the balancing parameters, and $a_1+a_2+a_3=1$.


\textbf{Statistical Constraint.} To further reduce the open space risk~\cite{toward}, we force latent features from different classes to approximate different multivariate Gaussian distributions $N(\bm{z};\bm{\mu}_k,\bm{I})$ by using the following KL-divergence loss function proposed in~\cite{cgdl}: 
\begin{equation}
\begin{aligned}
&\minus D_{KL}\big(p(\bm{z}|\bm{x},k) \ || \ q^{(k)}(\bm{z})\big)\\
&=\frac{1}{2}\sum_{j=1}^{J}\big(1+\log(\sigma_{j}^{2})-(\mu_j-\mu_j^{(k)})^2-\sigma_{j}^{2}\big)
\label{kl_loss}
\end{aligned}
\end{equation}
where $p(\bm{z}|\bm{x},k)$ is the class conditional posterior distributions and $q^{(k)}(\bm{z})$ is $k$-th multivariate Gaussian model (for more details please refer to~\cite{cgdl}). A fully-connected layer is adopted to map the one-hot encoding vector of the input's label to the latent space, and the mean of $k$-th Gaussian distribution $\bm{\mu}_k$ is obtained. A two-dimensional tSNE~\cite{tsne} visualization of latent feature distributions of M2IOSR on the CIFAR10 dataset~\cite{cifar10} is shown at the bottom-left of Fig.~\ref{framework}. It is observed that latent features are clustered into 10 multivariate Gaussian distributions and each distribution represents one known class.


\subsection{Training and Testing}
\textbf{Training.} There are two phases during the M2IOSR training procedure: the \emph{max-min} phase and the \emph{classification} phase. During the \emph{max-min} phase, following the loss function defined in Eqn.~\ref{loss_sum2}, the encoder and discriminators are trained jointly to maximize MI on positive pairs and minimize MI on negative pairs, as well as constrain the latent features to have desired statistical characteristics. During the \emph{classification} phase, the encoder is trained to minimize the cross-entropy loss between predicted scores and ground-truth labels.

\textbf{Testing.} A given image is passed through the encoder network to obtain the probability score vector $\bm{V}$ on known classes. Following the set-up in~\cite{gdfr}, the given image will be identified as an unknown sample if its maximum predicted probability $max(\bm{V})$ is smaller than a predetermined threshold $\tau$.
\section{Experiments}
\subsection{Implementation Details}
\label{imp}
To have a fair comparison with previous works, following~\cite{crosr}, we employ the VGG-13 as our backbone, please refer to~\cite{crosr} for details. We set the dimension of latent features to 32. The proposed M2IOSR is trained end-to-end by an SGD optimizer with an initial learning rate of 0.01 (multiply by 0.1 for every 50 epochs) and a momentum of 0.9. Batch-size is set to 64. For the loss function defined in Eqn.~\ref{loss_sum2}, $\beta_1$ equals to 0.5, $\beta_2$ equals to 1.0, and $\gamma$ equals to 0.1. For the local MI loss defined in Eqn.~\ref{local_loss}, $a_1$ equals to 0.7, $a_2$ equals to 0.1, and $a_3$ equals to 0.2. Networks trained by the proposed loss function should have comparable closed set testing accuracies with those trained only by the classification loss. The testing results of the closed set on each benchmark are shown in Table~\ref{close}. The decision threshold $\tau$ is fixed to 0.95.

\begin{table}
\begin{center}
\caption{Comparison of testing accuracies on closed sets between vanilla CNN models and our proposed approach M2IOSR. The training objective of M2IOSR is to maximize mutual information as well as classify known classes, but there is no large decline in testing accuracies on closed sets.}
\label{close}
\begin{tabular}{c c c c}
\hline
Architecture  & MNIST & SVHN & CIFAR10\\
\hline\hline
Vanilla CNN & 0.995 & 0.940 & 0.906\\
M2IOSR (ours) & 0.995 & 0.935 & 0.903\\
\hline
\end{tabular}
\end{center}
\end{table}


\subsection{Ablation Analysis}
\label{ablation_sec}
In this section, we analyze the effectiveness of each module used in the proposed method on the CIFAR10~\cite{cifar10} and CIFAR100 dataset~\cite{cifar100}. All classes in the CIFAR10 dataset are selected as known classes. Unknown classes are sampled from the CIFAR100 dataset varying from 10 to 100 (10, 14, 19, 25, 32, 42, 54, 71, 100), which means Openness level~\cite{toward} varies from 18\% to 59\%. The performance is measured by the macro-average F1-scores~\cite{f1} in all known classes and the unknown class against varying Openness levels.

We conduct experiments on the following baselines for ablation analysis. In all ablation experiments, the re-designed VGG-13 network defined in~\cite{crosr} is selected as the backbone, and if the maximum predicted score of input is less than the threshold $\tau$ defined in~\ref{imp}, this input will be classified as unknown. 

\textbf{I. CNN}: A vanilla convolutional neural network (CNN) with a closed set classifier is trained for known classification. During testing, the classifier will give out the probability scores of known classes.

\textbf{II. AE}: An auto-encoder (AE) model is set up for known classification, with a closed set classifier on the top of learned latent features. 

\textbf{III. DIM}: A deep infomax (DIM)~\cite{dim} model is built up for ablation analysis, where the setting is nearly the same as the original DIM method: only the $f_{16}(\bm{x})$ feature map and $f_{1}(\bm{x})$ feature vector are used for mutual information maximization, and the latent features are constrained to have desired independence characteristics through adversarial learning. A closed set classifier is on the top of learned latent features, which is trained for known classification. 

\textbf{IV. DIM+KL}: The model architecture and mutual information maximization procedure are the same as baseline III, but differently, the latent features are constrained to have conditional Gaussian distributions through the KL-divergence loss function proposed in~\cite{cgdl} (Eqn.~\ref{kl_loss}).

\textbf{V. DIM+KL+L1t4}: The model architecture and statistical constraint procedure are the same as baseline IV, but $l_{1t4}$ defined in Eqn.~\ref{local_loss} is added to the local MI loss function. $a_1$ and $a_2$ are set equal to 0.7 and 0.3 respectively (as $l_{4t4}$ is not used here, $a_3$ equals to 0).

\textbf{VI. DIM+KL+L4t4}: The model architecture and statistical constraint procedure are the same as baseline IV, but $l_{4t4}$ defined in Eqn.~\ref{local_loss} is added to the local MI loss function. $a_1$ and $a_3$ are set equal to 0.7 and 0.3 respectively (as $l_{1t4}$ is not used here, $a_2$ equals to 0).

\begin{figure} [htbp]
    \centering
    \includegraphics[scale=0.40]{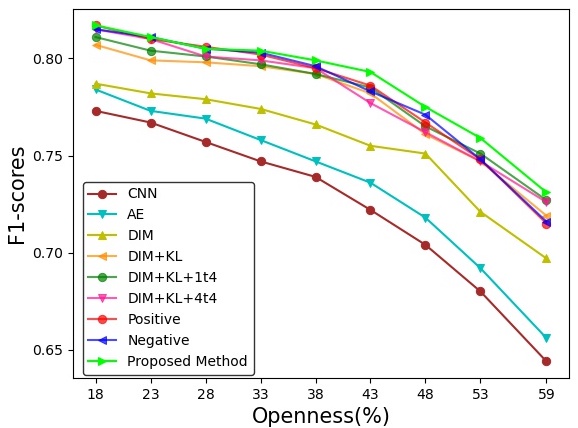}
    \caption{F1-scores against varying Openness levels in ablative experiments. The proposed method outperforms other baselines and achieves the highest F1-scores on all Openness levels.}
    \label{abla}
\end{figure}

\textbf{VII. Proposed Method}: The model architecture and statistical constraint procedure are the same as baseline IV, but $l_{1t4}$ and $l_{4t4}$ are both added to the local MI loss function as Eqn.~\ref{local_loss}. $a_1$, $a_2$ and $a_3$ are set equal to 0.7, 0.1 and 0.2 respectively. Besides, we also conduct ablative experiments when only using positive or negative pairs in the proposed method (namely \textbf{Positive} and \textbf{Negative} respectively in Fig.~\ref{abla}).

\begin{table*}
\begin{center}
\caption{Averaged F1-scores and their standard deviations in five randomized trials.}
\vspace{-2mm}
\label{state1}
\begin{tabular}{l c c c c c c}
\hline
Method & MNIST & SVHN & CIFAR10 & CIFAR+10 & CIFAR+50\\
\hline
\hline
Softmax & 0.768 $\pm$ 0.008 & 0.725 $\pm$ 0.012 & 0.600 $\pm$ 0.037 & 0.701 $\pm$ 0.012 & 0.637 $\pm$ 0.008 \\
Openmax~\cite{openmax} & 0.798 $\pm$ 0.018 & 0.737 $\pm$ 0.011 & 0.623 $\pm$ 0.038 & 0.731 $\pm$ 0.022 & 0.676 $\pm$ 0.026 \\
CROSR~\cite{crosr} & 0.803 $\pm$ 0.013 & 0.753 $\pm$ 0.019 & 0.668 $\pm$ 0.013 & 0.769 $\pm$ 0.016 & 0.684 $\pm$ 0.005 \\
CGDL~\cite{cgdl} & {0.837} $\pm$ {0.015} & {0.776} $\pm$ {0.010} & {0.655} $\pm$ {0.023} & {0.760} $\pm$ {0.024} & {0.695} $\pm$ {0.013} \\
GDFR~\cite{gdfr} & 0.821 $\pm$ 0.021 & 0.716 $\pm$ 0.010 & 0.700 $\pm$ 0.024 & 0.776 $\pm$ 0.003 & 0.683 $\pm$ 0.023 \\
\hline
M2IOSR & \textbf{0.892} $\pm$ \textbf{0.015} & \textbf{0.788} $\pm$ \textbf{0.012} & \textbf{0.733} $\pm$ \textbf{0.013} & \textbf{0.796} $\pm$ \textbf{0.008} & \textbf{0.744} $\pm$ \textbf{0.005}\\
\hline
\end{tabular}
\end{center}
\vspace{-0.16cm}
\end{table*}

\begin{table}
\begin{center}
\caption{F1-scores on the CIFAR10 dataset with four other datasets added to the testing set as unknown.}
\vspace{-2mm}
\label{state2}
\begin{tabular}{l c c c c}
\hline
{Method} & \tabincell{c}{ImageNet\\ Crop} & \tabincell{c}{ImageNet\\ Resize} & \tabincell{c}{LSUN\\ Crop} & \tabincell{c}{LSUN\\ Resize} \\
\hline\hline
Softmax & 0.639 & 0.653 & 0.642 & 0.647 \\
Openmax~\cite{openmax} & 0.660 & 0.684 & 0.657 & 0.668 \\
CROSR~\cite{crosr} & 0.721 & 0.735 & 0.720 & 0.749\\
C2AE~\cite{c2ae} & 0.837 & 0.826 & 0.783 & 0.801\\
CGDL~\cite{cgdl} & \textbf{0.840} & {0.832} & {0.806} & {0.812}\\
GDFR~\cite{gdfr} & {0.757} & {0.792} & {0.751} & {0.805}\\
\hline
M2IOSR & \textbf{0.840} & \textbf{0.834} & \textbf{0.818} & \textbf{0.826}\\
\hline
\end{tabular}
\end{center}
\vspace{-0.16cm}
\end{table}

The ablative results are drawn in Fig.~\ref{abla}. The vanilla CNN model (baseline I) gives the worst performance, and the AE architecture (baseline II) only improves the performance a little, which shows the reconstruction strategy is indeed helpful for OSR but the effectiveness is limited. The overall performance in these two baselines is not so good as the F1-scores drop dramatically with Openness levels increasing. In baseline III, this trend is alleviated and the performance is increased by introducing MI maximization strategy, which shows the effectiveness of this strategy for OSR. In baseline IV, there is an obvious performance improvement when latent features are constrained to class conditional Gaussian distributions by the KL-divergence loss function defined in~Eqn.~\ref{kl_loss}. The overall performance is improved a little by maximizing local MI on the pair $[f_{4}(\bm{x}),f_1(\bm{x})]$ (baseline V) or $[f_{4}(\bm{x}),f_4(\bm{x})]$ (baseline VI). As a result, the proposed method (baseline VII) outperforms other baselines and achieves the highest F1-scores on all Openness levels. Noted that if we only use positive or negative pairs in the proposed method, the performance is comparable with the proposed method when the Openness level is small (less than 38\%), but the performance degrades dramatically when the Openness rises to more than 38\%.

\begin{figure}[t]
\centering
\begin{tabular}{c@{\hspace{2mm}}c}
\includegraphics[width=0.45\linewidth]{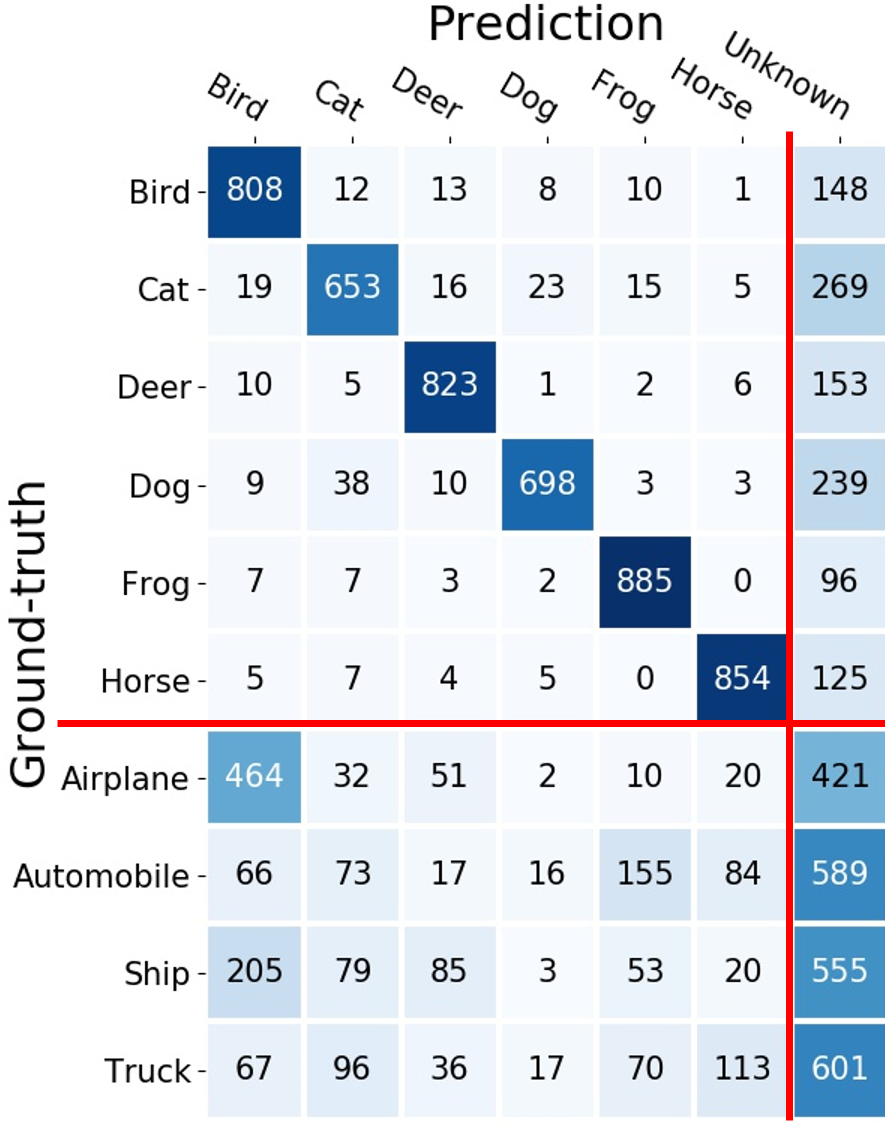}&
\includegraphics[width=0.45\linewidth]{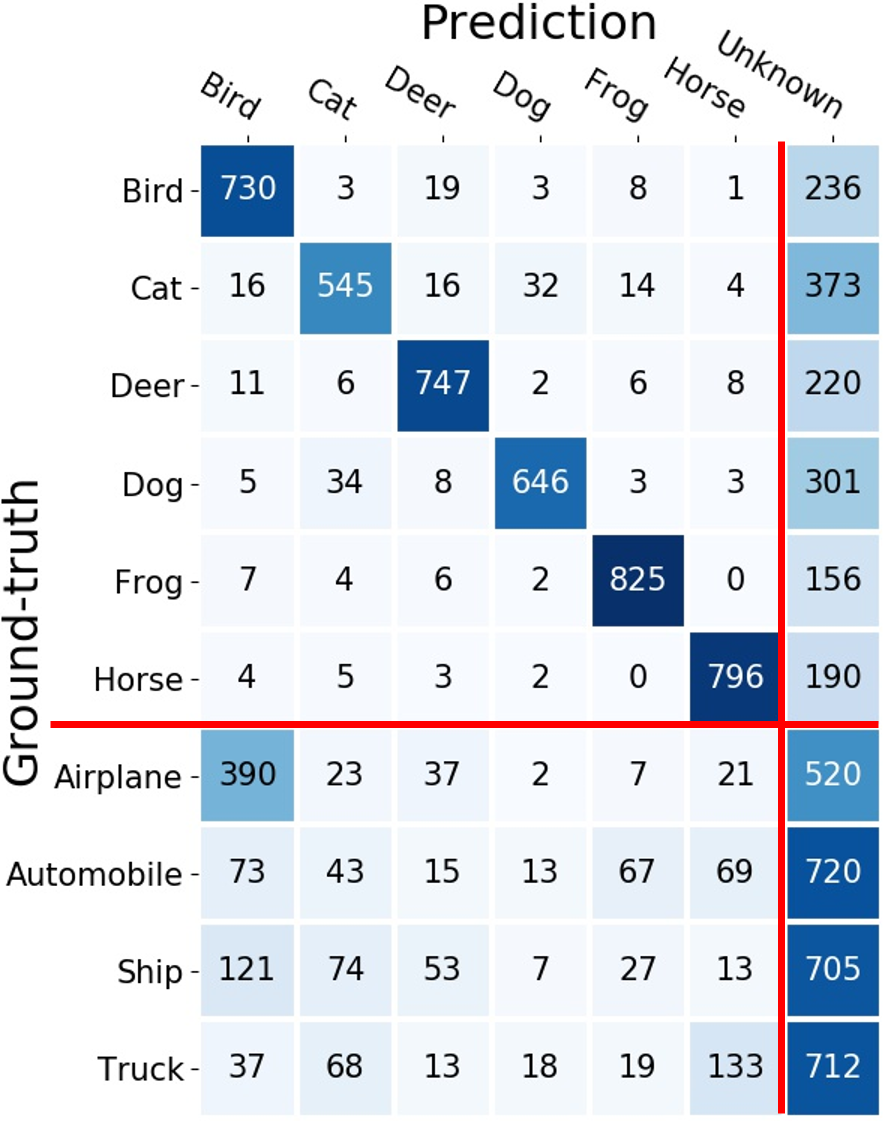}
\vspace{-0.1cm}\\
{\small\hspace{0.75cm}Auto-encoders} & {\small\hspace{0.75cm}Ours}
\end{tabular}
\vspace{-0.16cm}
\caption{\textbf{Confusion Matrix.} Here we show the confusion matrices produced by the AE model and our approach. The vehicle classes (airplane, automobile, ship, and truck) form the unknown class set and are not used in the training stage. Our approach performs much better on detecting unknown samples (the bottom-right part) than the AE model and significantly reduces errors in classifying unknown samples into known categories (the bottom-left part).}
\vspace{-0.36cm}
\label{cm}
\end{figure}

\subsection{Case Study}
\label{case_sec}
\xin{For heatmap drawing, we train an AE model and an M2IOSR model on the Dogs vs. Cats dataset, with the backbone of the VGG-19 network~\cite{vgg} (the last layer before latent features is replaced by a global pooling layer for Class Activation Mapping~\cite{cam}). The Dogs vs. Cats dataset has a larger image size (around $224 \times 224$) than CIFAR10 or CIFAR100 dataset ($32 \times 32$) and can generate much clearer heatmaps for comparison. Some visualizations of produced class activation maps are shown in Fig.~\ref{heatmap} (more examples are shown in supplementary materials). Compared with AE, the proposed method is more accurate in class-specific feature extraction and robust in target objects localization.}

We conduct another case-study on the CIFAR-10 dataset~\cite{cifar10} where all animal classes (bird, cat, deer, dog, frog, and horse) are selected as known classes, and the unknown class set consists of four vehicle classes (airplane, automobile, ship, and truck). The confusion matrices produced by the AE model and the proposed method are shown in Fig.~\ref{cm}. We divide each matrix into four parts by the red lines. It is observed that the results on known classes (the top-left part) are comparable, but the proposed method significantly increases the performance on detecting unknown samples (the bottom-right part) and reduces errors in classifying unknown samples into known categories (the bottom-left part). Fig.~\ref{cm} shows that our method can detect unknown samples more effectively than the AE model and maintain high accuracy in known classification.

\subsection{Comparison with State-of-the-art Results}
Here, we report our results on benchmarks and compare the proposed M2IOSR with existing OSR methods. An ideal OSR classifier should accurately classify known classes and effectively identify unknown samples, therefore in all experiments, the results are measured by macro-averaged F1-scores among all known classes and the unknown class. 

The results of our M2IOSR are demonstrated in Table~\ref{state1}, \ref{state2} and \ref{state3}. In Table~\ref{state1}, we conduct experiments on four standard image datasets: MNIST~\cite{mnist}, SVHN~\cite{svhn}, CIFAR10~\cite{cifar10} and CIFAR100~\cite{cifar100}. For each of the MNIST, SVHN, and CIFAR10 datasets, known classes consist of six randomly selected categories, and the remaining four categories constitute the unknown class. Meanwhile, in the CIFAR+10 /CIFAR+50 task, 4 vehicle categories in the CIFAR10 dataset (airplane, automobile, ship, and truck) are selected as known classes, and 10/50 animal categories in the CIFAR100 dataset are randomly selected to constitute the unknown class. The results shown in Table~\ref{state1} are averaged F1-scores and their standard deviations among 5 runs on different known and unknown separations. \xin{Noted that some previous works used Area under ROC curve (AUROC) to measure the performance on these tasks, here we used released codes to evaluate OSR performance on macro-averaged F1-scores and report these results in Table~\ref{state1}.} As shown in this table, the proposed approach significantly outperforms existing OSR methods and achieves the highest averaged F1-scores.

\begin{table}
\begin{center}
\caption{F1-scores on the MNIST dataset with three other datasets added to the testing set as unknown.}
\vspace{-0.16cm}
\label{state3}
\begin{tabular}{l c c c c}
\hline
Method & Omniglot & MNIST-Noise & Noise \\
\hline\hline
Softmax & 0.595 & 0.801 & 0.829\\
Openmax~\cite{openmax} & 0.780 & 0.816 & 0.826\\
CROSR~\cite{crosr} & 0.793 & 0.827 & 0.826\\
CGDL~\cite{cgdl} & {0.850} & {0.887} & {0.859}\\
\hline
M2IOSR & \textbf{0.924} & \textbf{0.923} & \textbf{0.929}\\
\hline
\end{tabular}
\end{center}
\vspace{-0.36cm}
\end{table}

In Table~\ref{state2}, following the protocol defined in \cite{crosr}, the CIFAR10 dataset consist of known classes, and two other datasets (ImageNet \cite{imagenet} and LSUN \cite{lsun}) are respectively selected as the unknown class. To make the unknown sample size the same as known samples, we crop or resize unknown samples to $32 \times 32$, then ImageNet-resize, ImageNet-crop, LSUN-resize, and LSUN-crop are generated. In each dataset, there are a total number of 10,000 testing images, which is the same as the CIFAR10 testing set. This makes the known-to-unknown ratio 1:1 during testing. The F1-scores of the proposed method and other OSR methods are shown in Table~\ref{state2}. We can see that the proposed method achieves new state-of-the-art performance. Although the proposed method improves the performance a little on the ImageNet dataset, the improvement on the LSUN dataset is significant.

In another experiment, following the set-up in~\cite{crosr}, the complete MNIST training set consists of all known classes. Three other datasets, Omniglot~\cite{omniglot}, MNIST-Noise, and Noise are respectively selected as the unknown class.
Omniglot is a dataset containing various alphabetic characters. Noise is a synthetic dataset, and each pixel value of its sample is independently from a uniform distribution between 0 and 1. MNIST-Noise is also a synthesized dataset by adding noise to the MNIST testing samples. Each unknown dataset has the same number of samples as the MNIST testing set, making the ratio of the known-to-unknown is 1:1. The F1-scores on these datasets are listed in Table~\ref{state3}. Our method achieves new state-of-the-art performance on all given tasks.

\begin{table}
\begin{center}
\caption{Comparison of running time and number of parameters. All the models are based on VGG-13.}
\vspace{-0.16cm}
\label{time}
\begin{tabular}{l c c c c}
\hline
Method &&  Time & Parameter\\
\hline\hline
Softmax && 0.179 & 9,428k \\
Openmax~\cite{openmax} && 0.181 & 9,428k \\
CROSR~\cite{crosr}  && 0.184 & 10,003k\\
CGDL~\cite{cgdl} && 0.996 & 291,352 k\\
GDFR~\cite{gdfr} && 0.772 & 10,424 k\\
\hline
M2IOSR (ours) && 0.618 & 9,444 k\\
\hline
\end{tabular}
\end{center}
\vspace{-0.36cm}
\end{table}

\xin{In Table~\ref{time}, we measure the running time (milliseconds/image) of feature extraction and the number of parameters of each model when testing the CIFAR10 dataset on a single Tesla V100 graphic processor. Compared with other OSR methods, the proposed method has a relatively small number of parameters. However, as we introduce positive/negative pairs and global/local MI maximization strategy, the highest F1-scores are achieved at a reasonable cost of longer running time.}

\section{Conclusion}
In this work, we propose an OSR approach termed Maximal Mutual Information Open Set Recognition (M2IOSR). We analyze the pixel-level reconstruction strategy used in most previous OSR methods and find that this strategy is commonly over-demanding for OSR tasks and makes class-specific features contained in target objects less pronounced. To address these shortcomings, we discard the reconstruction strategy and propose an MI-based method with a streamlined architecture. The proposed M2IOSR only employs an encoder for class-specific feature extraction, thus is much more lightweight than previous methods with a decoder. Our multi-scaled MI maximization strategy encourages the encoder to establish strong interdependence between the input image and its latent features, which greatly improve the model's ability to classify known classes and detect unknown samples. Moreover, to further reduce the open space risk, latent features are constrained to class conditional Gaussian distributions by a KL-divergence loss function. The proposed M2IOSR greatly enhances the performance of our baseline and leads to new state-of-the-art results on several standard benchmarks. 

{\small
\bibliographystyle{ieee_fullname}
\bibliography{egbib}
}

\end{document}